\documentclass[11pt,a4paper]{article}
\usepackage[T1]{fontenc}
\usepackage[utf8]{inputenc}
\usepackage{lmodern}
\usepackage[margin=1in]{geometry}
\usepackage{microtype}
\usepackage{amsmath,amssymb,amsthm}
\usepackage{graphicx}
\graphicspath{{fig/}}
\usepackage{booktabs,tabularx,array,longtable}
\usepackage[numbers,sort&compress]{natbib}
\usepackage{fancyhdr}
\usepackage[hidelinks]{hyperref}
\hypersetup{pdftitle={Attending to Memory Writes Reduces Seed Variability in Recurrent Length Generalization},
  pdfsubject={Update-level attention, recurrent learning, and length generalization}}
\fancypagestyle{firstpage}{%
  \fancyhf{}

  \fancyfoot[C]{\footnotesize\url{https://github.com/swgoo/trace_relay}\\[2pt]\normalsize\thepage}
}

\newcommand{\model}{TraceRelay}

\newcommand{\attn}{\operatorname{Attn}}

\newcommand{\paperfigure}[1]{\includegraphics[width=\linewidth]{#1}}
\title{Attending to Memory Writes Reduces Seed Variability in Recurrent Length Generalization}
\author{%
  Sungwoo Goo$^{1}$ \qquad Hwi-yeol Yun$^{1}$ \qquad Sangkeun Jung$^{2}$\\
  \small $^{1}$College of Pharmacy, Chungnam National University\\
  \small $^{2}$Department of Computer Science \& Engineering, Chungnam National University\\
  \small Daejeon, Republic of Korea\\
  \small \texttt{swgoo91@gmail.com, hyyun@cnu.ac.kr, hugmanskj@gmail.com}%
}
\date{}

\begin{document}
\maketitle
\thispagestyle{firstpage}
\begin{abstract}
Recurrent memory requires learning useful updates, but attention added to
assist learning can also compromise length generalization. We study
\emph{bounded delta attention}: causal attention over uncorrected update
proposals whose bounded correction enters the recurrent update. Its
implementation in \model\ combines additive phase recurrence with attention
computed before a parallel scan. On prefix-supervised modulo-5, trained at
lengths 1--40, three findings characterize its behavior at length 80. First,
upper delta attention attains $70.81\pm8.26$\% accuracy across seven seeds,
versus $69.21\pm22.42$\% for the local-attention recurrent baseline
(mean $\pm$ sample SD): a 63.2\% reduction in SD with a similar mean.
Upper state attention also reduces variation, reaching $74.75\pm6.24$\%,
so this benefit is not unique to update-level attention. Disabling the
upper delta correction retains $70.62\pm8.60$\%. Second, blocking gradients
at the delta-attention inputs changes accuracy to $65.25\pm12.40$\% in the
matched seven-seed comparison, increasing SD by 50.2\%. It also delays a
99\% validation threshold on average among the same six seeds that reach it;
the 95\% threshold is reached earlier on average with input-gradient blocking across all seven seeds.
Third, lower delta attention exhibits substantially less
extrapolation degradation than lower state attention: 62.60\% versus
19.40\%, against a 71.06\% baseline. Removing its correction reduces
accuracy to 22.56\%; its advantage over lower state attention also persists
under input-gradient blocking. These observations motivate bounded
update-level attention as a recurrent learning scaffold, combining reduced
seed variation with less severe degradation in a placement where state
attention fails. Improved credit assignment is a candidate explanation,
not an established mechanism. Longer-length failures and mixed all-layer
results delimit the evidence; next-token-trained recurrent memory remains
a future objective.

\end{abstract}

\section{Introduction}
\label{sec:introduction}
Maintaining useful information over a long interaction requires more than recovering an earlier token.
A model may need to revise a belief, accumulate a quantity, or preserve the current status of a task as new observations arrive.
Such computations motivate recurrent state updates, but access to a recurrent path does not ensure that training consistently discovers an update rule that generalizes.
State-tracking experiments show that strong performance within the training lengths can coexist with length-specific solutions and poor transfer to longer sequences \citep{ebrahimi2026induction}.

Attention offers a natural way to connect distant computations during learning.
Temporal attention and skip connections have previously improved recurrent optimization and credit assignment \citep{ke2018sab,soo2023dasc}.
However, attention can also change which computation a recurrent hybrid learns.
Delayed Attention Training (DAT) demonstrates that jointly trained attention can compromise an RNN's length generalization and that delaying attention can improve it \citep{phan2025delayed}.
SWAX shows that shorter training-time attention windows can improve long-context recall in recurrent hybrids \citep{cabannes2025swax}.
These findings motivate examining how a temporal shortcut is coupled to recurrence, alongside when it is trained and how far it can reach.

We study a particularly simple coupling: attention over proposed recurrent increments.
A local computation first produces an update $\delta_t$.
Causal attention accesses the current and previous uncorrected updates and produces a bounded correction.
The corrected increment then enters the recurrent transition.
We call this operation \emph{bounded delta attention}.
In the \model\ implementation used here, the transition accumulates phase angles and exposes their sine and cosine features.
The correction can therefore be bounded directly in radians.

Our empirical question is whether this update-level route can support useful recurrent learning while mitigating the extrapolation degradation observed with long-window attention over state representations.
We use modulo-5 prefix tracking, which has a small sufficient state and an exactly repeatable update rule.
Every input position receives its prefix-state label; no generated intermediate answer is fed back as an input. This isolates a state-update problem while providing dense supervision.

The results center on three observations.
First, upper layer's delta attention reduces the observed spread across seven seeds while retaining a similar
mean to the local-attention recurrent baseline; most performance remains when its correction is disabled at evaluation.
Second, blocking the attention-input gradient route increases the length-80 spread again in a matched seven-seed comparison. It delays the 99\% validation-accuracy threshold on average among the same six seeds that reach it, while reaching the 95\% threshold earlier on average across all seven seeds.
Third, lower layer's delta attention causes substantially less extrapolation degradation than state attention, but depends strongly on its correction at inference.
This advantage over lower state attention also persists under input-gradient blocking, indicating that the forward correction contributes alongside the additional gradient route.

Upper layer's state attention remains a strong control, with a higher length-80 mean and smaller SD than upper delta attention.
Scheduled removal, training-time diagnostics, and all-layer variants further characterize the intervention behavior.
The contribution is therefore a specific bounded update mechanism and evidence about how it affects learned variability and inference-time dependence, rather than a general ranking of attention types.

The evidence also bounds the interpretation.
The attention targets, parameter counts, direct receptive fields, and output constraints differ between the state- and delta-attention variants.

\section{Related work}
\label{sec:related_work}

\paragraph{Attention and temporal shortcuts for recurrent learning.}
Attention has long complemented recurrent computation.
\citet{bahdanau2015align} let a recurrent translation decoder attend to source representations, reducing reliance on a single fixed-length encoding.
Sparse Attentive Backtracking (SAB) attends to past hidden states to create selective temporal paths for credit assignment, and evaluates generalization to longer sequences \citep{ke2018sab}.
Dynamics-aligned skip connections (DASC) assist optimization without attention by combining coarse and fine recurrent trajectories and annealing the auxiliary connections away \citep{soo2023dasc}.
These methods establish temporal shortcuts and removable training aids as useful approaches to recurrent learning.

Our focus is the signal accessed by attention and its entry point into recurrence.
The proposed branch reads \emph{raw update proposals}, applies a bounded correction, and supplies the corrected increments to a fixed additive phase transition.
Because the proposals are independent of the same layer's accumulated phase, attention can be evaluated before the scan.
This organization preserves parallel evaluation while introducing content-dependent paths among proposed updates.

\paragraph{Attention can also compete with recurrent learning.}
Delayed Attention Training (DAT) improves hybrid length generalization by postponing attention training;
its modulo-only experiments also show failure where a pure LSTM generalizes \citep{phan2025delayed}.
SWAX finds that wider sliding windows can weaken long-term memorization in attention--xLSTM hybrids and studies stochastic windows to balance local access and recurrent memory \citep{cabannes2025swax}.
\citet{irie2025blending} further show that synchronous versus delayed delivery to local-attention/DeltaNet memory systems changes algorithmic performance.
Together, these results motivate examining attention's coupling to recurrence, alongside its training schedule and reach.
We study a bounded route into the update stream within this established problem.

\paragraph{Structured recurrence and phase representations.}
Unitary RNNs use complex-valued, norm-preserving recurrent matrices to address difficult long-term optimization \citep{arjovsky2016unitary,wisdom2016unitary}.
The Linear Recurrent Unit combines linear, diagonal recurrence with suitable initialization and normalization for efficient parallel training \citep{orvieto2023lru}.
Bounded delta attention extends the current implementation by correcting unaccumulated increments.
The phase representation makes the correction scale interpretable as a per-update angular bound, whose implications for optimization and extrapolation remain empirical.

\paragraph{State tracking as a controlled diagnostic.}
\citet{tang2026entities} analyze natural-language entity tracking and find evidence for query-dependent consolidation rather than incremental state maintenance in the studied pretrained models.
Their mechanistic findings motivate our concern about learning ongoing recurrent updates.
\citet{ebrahimi2026induction} find substantial differences between recurrent networks and Transformers in modular-state-tracking sample efficiency and cross-length transfer under multiple supervision regimes.
\citet{buitrago2025length} connect recurrent extrapolation failures to incomplete training coverage of attainable states.
These findings distinguish fitting training examples from learning reusable transitions.
Our modulo-five task provides aligned prefix-state supervision without previous-answer inputs, enabling controlled comparisons of the resulting update computations.
\section{Bounded attention over recurrent updates}
\label{sec:method}

We study where a recurrent model should use a wider attention window: over representations containing accumulated state, or over the increments that update that state.
TraceRelay provides both a local attention pathway and an explicit phase recurrence in every layer.
A window of $W$ always includes the previous $W$ positions and the current position.
All attention operations are causal multi-head attention with RoPE and learned query, key, value, and output projections.

% Generated by scripts/build_artifacts.py.
\begin{figure}[tb]
\centering
\paperfigure{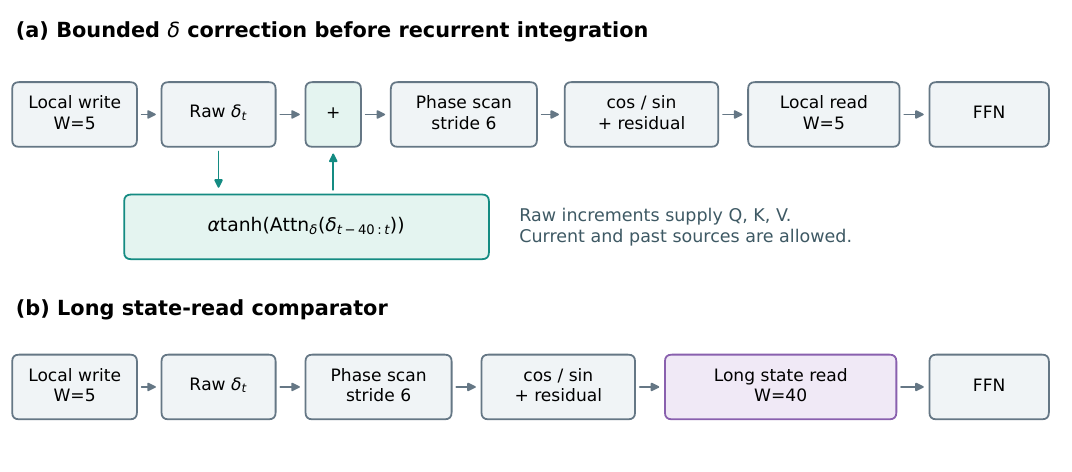}
\caption{Two locations for a long-window attention operation within a TraceRelay layer. (a) $\delta$ attention reads raw local phase increments, adds the coordinate-wise bounded correction $\alpha\tanh(\operatorname{Attn}_{\delta}(\delta_{t-40:t}))$, and then integrates the corrected increment through the stride-six recurrent phase update. The long correction uses neither accumulated phase states nor an independent memory controller. (b) The state-read comparator widens read attention after the recurrent update and state-feature construction. The diagram omits projection and normalization details given in the text. A window of 40 includes 40 past positions and the current position; the correction branch is enabled only in the designated layers.}
\label{fig:architecture}
\end{figure}

\subsection{Local writes, phase inheritance, and reads}
\label{sec:trace-relay}

Let $h_t\in\mathbb{R}^{d}$ denote the input to one layer, suppressing the layer index.
The separate learned RMSNorm layers $N_w$, $N_r$, and $N_f$ are applied before write attention, read attention, and the SwiGLU feed-forward sublayer, respectively.
The operator $\operatorname{Attn}_{W}(z)_t$ attends to $z_{t-W:t}$.
A short write operation produces a raw phase increment:
\begin{align}
 a_t &= \operatorname{Attn}_{W_w}(N_w(h))_t, \\
 \delta_t &= \pi\tanh(P_{\delta}a_t).
 \label{eq:raw-delta}
\end{align}
Here $P_{\delta}\in\mathbb{R}^{d\times d}$ is a learned, bias-free linear projection applied to the write-attention output at each position, separately from the attention module's own output projection.
The coordinatewise $\tanh$ and factor $\pi$ convert its output into a raw phase increment bounded in magnitude by $\pi$ radians per coordinate.
The baseline applies $\widetilde\delta_t=\delta_t$. With $s=W_w+1$, phase inheritance is
\begin{equation}
 \theta_t= (\theta_{t-s}+\widetilde\delta_t) \mod 2\pi,
 \qquad \theta_t=0\text{ for }t\leq0.
 \label{eq:phase-scan}
\end{equation}
This defines $s$ interleaved recurrent lanes. The write at $t$ attends to the current local interval and immediately updates the phase inherited from just before that interval.
Write queries, keys, and values depend only on this layer's inputs; a strided prefix scan therefore computes the phases in parallel.

The phase enters the residual stream before read attention:
\begin{align}
 m_t &= h_t+W_{\phi}[\cos\theta_t;\sin\theta_t], \\
 u_t &= m_t+P_r\operatorname{Attn}_{W_r}(N_r(m))_t, \\
 h'_t &= u_t+\operatorname{SwiGLU}(N_f(u_t)).
 \label{eq:state-read}
\end{align}
The learned projection $P_r$ follows the attention output projection. The phase feature map is coordinatewise; the equations omit masking, under which uninitialized memory contributes zero.

\subsection{Attention over raw increments}
\label{sec:delta-attention}

An optional attention module computes a bounded correction before the scan:
\begin{align}
 c_t &= \operatorname{Attn}_{W_{\delta}}(\delta)_t, \\
 \epsilon_t &= \alpha\tanh(c_t), \\
 \widetilde\delta_t &= \delta_t+\epsilon_t.
 \label{eq:delta-correction}
\end{align}
Its queries, keys, and values are projections of \emph{raw} increments, without an additional RMSNorm.
Corrected increments are never fed back as this attention module's source.
Thus the operation can connect updates across recurrent lanes while retaining parallel computation of corrections and the subsequent scan.

The nonnegative coefficient $\alpha$ bounds each correction coordinate in radians: $|\epsilon_{t,j}|\leq\alpha$.
Holding the incoming phase and raw increment fixed, the associated phasor perturbation obeys
\begin{equation}
 |e^{i(\theta+\delta+\epsilon)}-e^{i(\theta+\delta)}|
 =2|\sin(\epsilon/2)|\leq|\epsilon|\leq\alpha.
 \label{eq:phasor-bound}
\end{equation}
This is a per-update, per-coordinate bound.

\subsection{State-attention controls and direct reach}
\label{sec:controls}

The baseline has $W_w=W_r=5$ in all three layers, hence $s=6$.
A \emph{state-attention} control widens one existing read window to 40.
A single-layer $\delta$-attention model instead keeps all reads short and adds an increment-attention module with $W_{\delta}=40$.
Lower, middle, and upper placements refer to layers one, two, and three.
Widening a read does not change the same layer's phase recurrence for fixed layer inputs, but it changes the layer output and can consequently affect upper layers' writes and future recurrent states.
The wider window covers every training prefix but remains fixed at evaluation;
it is not unrestricted full attention on longer sequences.

Removing only the phase-inheritance edges leaves direct receptive-field width
\begin{equation}
 R_{\mathrm{direct}}=1+\sum_{\ell=1}^{3}
  (W_{w,\ell}+W_{r,\ell}+\overline W_{\delta,\ell}),
 \label{eq:direct-field}
\end{equation}
where $\overline W_{\delta,\ell}=0$ when no correction contributes.
The baseline, single state-attention, single $\delta$-attention, and all-layer $\delta$-attention models have widths 31, 66, 71, and 151, respectively, when $\alpha>0$. They have different parameter allocations and direct reach (Table~\ref{tab:controls}); this is not an exact capacity-matched substitution.
In particular, length 80 lies inside the all-layer model's direct field, so its accuracy there alone does not establish reliance on inherited phase beyond local access.

\section{Experiments}
\label{sec:experiments}

\subsection{Modular state tracking and training}

Inputs are IID symbols $x_t\sim\operatorname{Uniform}\{0,\ldots,4\}$, with target state
\begin{equation}
 y_0=0,\qquad y_t=(y_{t-1}+x_t)\bmod5.
 \label{eq:task}
\end{equation}
A bias-free projection embeds each one-hot symbol, and a final normalization and classifier predict the inclusive prefix state. The loss is mean cross-entropy over all input-aligned positions in a minibatch. This aligned intermediate-state supervision (ACoT; \citealp{ebrahimi2026induction}) feeds neither targets nor previous predictions back into the input. It is not natural-language next-token prediction.

All reported training runs complete 40,000 updates. Each minibatch contains 64 fresh sequences of a single length sampled uniformly from $1$--$40$. Models have three layers, hidden width 128, feed-forward width 512, four attention heads, and zero dropout. AdamW uses learning rate $3\times10^{-4}$, weight decay 0.01, 100 warmup updates, cosine decay to one tenth of the peak, and gradient clipping at 1. Training uses BF16. Table~\ref{tab:protocol} records the remaining shared settings.

\begin{table}[tb]
\centering
\caption{Shared training and evaluation protocol. Main upper comparisons use seeds 42--48; supplementary placement, scale, and scaffold comparisons use seeds 42--44 for every condition, including the corresponding upper subsets.}
\label{tab:protocol}
\small
\begin{tabularx}{\linewidth}{@{}lX@{}}
\toprule
Component & Setting \\
\midrule
Task and supervision & Modulo-5 inclusive prefix state; one-hot symbols; label at every position; no target feedback \\
Architecture & Three layers; hidden width 128; SwiGLU width 512; four attention heads; zero dropout \\
Training data & Fresh IID symbols; one length $L\sim\mathrm{Uniform}\{1,\ldots,40\}$ per minibatch of 64 sequences \\
Training budget & 40,000 updates; 2,560,000 sequences per run \\
Optimizer & AdamW: learning rate $3\times10^{-4}$, weight decay 0.01, $\beta=(0.9,0.999)$, $\epsilon=10^{-8}$ \\
Schedule and numerics & 100 warmup updates; cosine decay to $0.1$ of peak; BF16; gradient norm clipped at 1 \\
Local and long windows & Write/read windows 5; selected state-read or increment-attention window 40; windows count past positions plus current \\
Validation & 2,000 length-40 sequences; ordinary cross-entropy; generator seed 20261023 \\
Checkpoints & \textsc{Last} at update 40,000 is primary; minimum ID-validation-CE \textsc{Best} is a sensitivity analysis \\
Evaluation & 1,024 fixed held-out sequences per length; generator seed 20261024; final-position accuracy \\
Common evaluation lengths & 20, 40, 80, 160, and 320; additional lengths only where recorded \\
\bottomrule
\end{tabularx}
\end{table}

\begin{table}[tb]
\centering
\caption{Model configurations and direct receptive fields with phase-inheritance edges excluded. Every layer retains write window 5 and phase stride 6. The correction scale is positive in the listed $\delta$ fields; setting it to zero restores direct field 31 while retaining the trained parameters.}
\label{tab:controls}
\small
\setlength{\tabcolsep}{4pt}
\begin{tabularx}{\linewidth}{@{}lXXrr@{}}
\toprule
Design & Read windows & $\delta$ windows & Parameters & $R_{\mathrm{direct}}$ \\
\midrule
Baseline & All 5 & None & 1,182,213 & 31 \\
State attention & One 40; others 5 & None & 1,182,213 & 66 \\
Single $\delta$ attention & All 5 & One 40 & 1,247,749 & 71 \\
All-layer $\delta$ attention & All 5 & All 40 & 1,378,821 & 151 \\
\bottomrule
\end{tabularx}
\par\smallskip
\begin{minipage}{\linewidth}
\footnotesize
Lower, middle, and upper placements have identical parameter counts within each attention design. State attention widens an existing reader; $\delta$ attention adds 65,536 parameters per active layer. The two designs therefore differ in parameter allocation, direct reach, attention inputs, and correction constraint.
\end{minipage}
\end{table}

\subsection{Comparisons and training seeds}

The main baseline, upper state attention, upper $\delta$ attention, and upper input-SG comparisons use the same seven seeds 42--48.
Supplementary placement, scale, and scaffold comparisons use the same seeds 42--44 for every condition, including the corresponding baseline and upper-layer subsets.
The principal correction scale is $\alpha=0.05$, selected after exploratory completed runs with $0.2$, $0.1$, and $0.05$.
The seven-seed comparisons test variability under the same task and protocol.

For a given seed, all shared parameters in the $\delta$ variants are copied from an identically initialized baseline. The additional attention weights use a separate seeded stream, with the same initial values at each tested placement and independent parameters when present in multiple layers. The post-initialization random-number state is restored, and symbol and length streams are matched. The state-attention extension uses the historical source snapshot; preflight checks verify matching baseline initialization, training settings, and a reproduced historical evaluation.

Additional all-layer experiments apply correction attention in every layer. Their three schedules are constant $\alpha=0.05$ (A), constant $0.02$ (B), and $0.05$ through update 24,000 followed by linear decay to zero at update 36,000 (C). Thus C trains its final 4,000 updates with zero correction and has direct field 31 at the final checkpoint, despite retaining the extra parameters. These experiments use seeds 42--44 and the same total budget.

\subsection{Training-path ablations and diagnostics}
\label{sec:scaffold-protocol}

We complete a further 12 runs: four configurations with seeds 42--44 and the
same 40,000-update budget. Upper fixed repeats the selected upper
$\alpha=0.05$ configuration with diagnostics. Upper input-SG and lower input-SG
stop gradients at all raw-update inputs to correction attention, while keeping
$\alpha=0.05$. Writing $\operatorname{sg}$ for stop-gradient, the intervention is
\begin{equation}
 \tilde\delta_t=\delta_t+\alpha\tanh\!\left(
 \attn_\delta(\operatorname{sg}(\delta_{t-W_\delta:t}))_t\right).
 \label{eq:input-sg}
\end{equation}
The notation includes the current query and all current/past key and value
inputs. It removes their additional gradient route through the helper,
not the main recurrent gradient through the uncorrected update. The attention
projections remain trainable and the correction still enters the carry.
For identical weights and inputs, input-SG leaves forward values unchanged;
parameter counts and direct receptive fields also match the full-gradient
configuration. It does not isolate past-value gradients from query gradients.

Upper decay retains all gradient paths and uses the same 24k--36k linear
schedule as the earlier all-layer C experiment, but only at the upper layer.
Its final 4,000 updates have zero correction. All four configurations match
the original initialization and training streams. The instrumented upper-fixed
runs reproduce the historical final weights and data/length generator states
bit-for-bit, and the upper-fixed and upper-decay checkpoints match at initialization and
24k. The repeated fixed runs therefore add measurements, not independent seeds.

The upper input-SG comparison uses seven distinct training seeds 42--48,
each completing 40,000 updates. Preflight checks confirm the corresponding
full-gradient initialization and post-initialization random-number state;
the model settings, training budget, and symbol/length streams are matched.
The input-gradient comparison counts each training seed once. Lower input-SG, upper
decay, and the four-condition diagnostic comparison retain seeds 42--44.

Diagnostics use 512 fixed validation examples at each of lengths 40 and 80,
with unchanged inputs for ON and OFF. We record 24 checkpoints: initialization,
updates 250, 500, and 1,000, then every 2,000 updates through 40,000. Measurements
include aligned cross-entropy and the coordinate/token RMS of the raw update
and actual scaled correction. The absolute ON--OFF cross-entropy difference
is computed within each seed before averaging. These probes neither select
checkpoints nor control training; their random-number states are restored.
They are distinct from the 1,024-example final test evaluations. As a
complementary learning-speed summary, we record the first ordinary 2,000-example
length-40 validation final-position accuracy of at least 99\%, checked every
250 updates.
For the seven-seed upper input-gradient comparison, we also compute the first
95\% crossing from the same logged validation evaluations.

\subsection{Checkpoint selection and evaluation}

The primary checkpoint is \textsc{Last}, after 40,000 updates. \textsc{Best}, chosen solely by minimum ordinary length-40 validation cross-entropy, provides a sensitivity analysis; extrapolation scores do not select checkpoints. Validation uses 2,000 fixed length-40 examples. Final evaluation uses 1,024 independently generated held-out sequences per length, shared across configurations and seeds and verified by fingerprints. The principal metric is final-position accuracy, with chance level 20\%. All seven-seed comparisons include lengths 20, 40, 80, 160, and 320; additional recorded lengths are reported only where available.

For $\delta$-\textsc{Off}, we reevaluate the same checkpoint with runtime $\alpha=0$ and unchanged inputs, without retraining. This probes reliance on the correction at inference, not equivalence of training trajectories to the baseline. An inference-time intervention is not an independently trained control. We report means, sample standard deviations, and individual seed outcomes descriptively; we do not treat \textsc{On}/\textsc{Off} evaluations of one checkpoint as additional training seeds.

\section{Results}
\label{sec:results}
We examined whether an attention route can assist recurrent learning while limiting the extrapolation degradation associated with a long state read.
Three findings organize the results: upper delta attention reduces observed seed variation, an input-gradient intervention partially reverses that reduction, and lower delta attention causes much less degradation than lower state attention while remaining necessary at inference.
Upper state attention also reduces seed variation and is included as a central control.
Unless marked BEST, results use the checkpoint after 40,000 optimizer updates.
Accuracy is measured at the final input position.
SDs are sample standard deviations across training seeds, in percentage points, not confidence intervals. Chance accuracy is 20\%.

\subsection{Upper attention reduces seed variation; the delta effect persists after removal}
\label{sec:seven-seed}
% Generated by scripts/build_artifacts.py.
\begin{figure}[tb]
\centering
\paperfigure{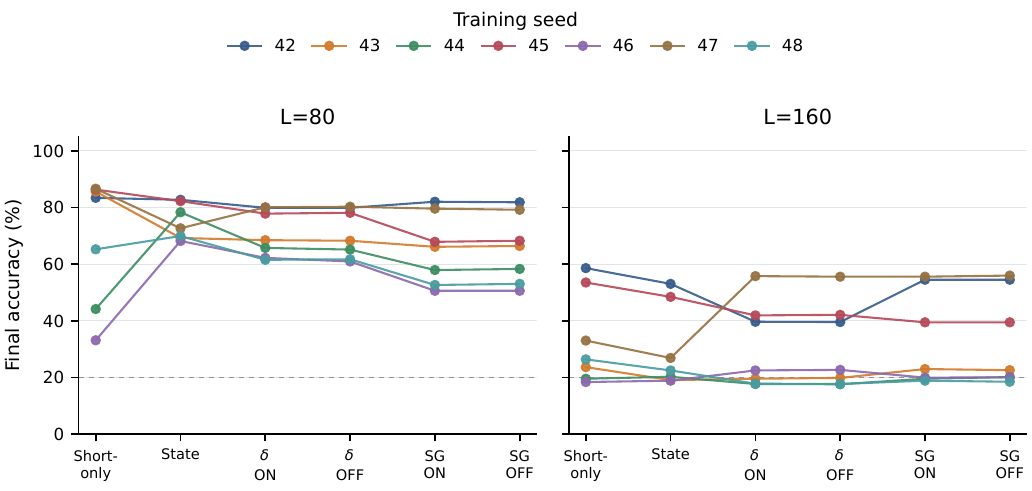}
\caption{Matched training-seed outcomes for the seven-seed upper-layer comparison at LAST (40k updates). Each line connects models with the same training seed. The $\delta$ models use $\alpha=0.05$, and OFF sets the correction scale to zero at inference on the same weights. SG blocks the raw-update Q/K/V input gradients through correction attention during training. The upper state comparator widens the last read window to 40. At $L=80$, upper state has both a higher mean and a lower sample standard deviation than upper $\delta$ ON (74.75\% $\pm$ 6.24 versus 70.81\% $\pm$ 8.26). Input-SG changes upper delta accuracy to 65.25\% $\pm$ 12.40. OFF closely follows ON within each training condition; performance degrades substantially at $L=160$.}
\label{fig:seven-seed}
\end{figure}

% Generated by scripts/build_artifacts.py.
\begin{table}[htbp]
\centering
\caption{Upper-layer comparison over seven seeds (42--48), LAST at 40k updates. Entries are final-position accuracy (\%), mean $\pm$ sample standard deviation. ON and OFF share weights; OFF disables only the $\delta$ correction at inference. Input-SG blocks the raw-update Q/K/V gradients through the correction branch. Upper state uses a longer read window and is trained separately.}
\label{tab:upper-seven}
\small
\setlength{\tabcolsep}{4pt}
\begin{tabular}{lrrrr}
\toprule
Condition & $L=40$ & $L=80$ & $L=160$ & $L=320$ \\
\midrule
Short-only & $98.76 \pm 0.63$ & $69.21 \pm 22.42$ & $33.29 \pm 16.34$ & $22.92 \pm 5.59$ \\
Upper state & $99.09 \pm 1.23$ & $74.75 \pm 6.24$ & $29.84 \pm 14.59$ & $23.03 \pm 5.82$ \\
Upper $\delta$ ON & $99.16 \pm 0.42$ & $70.81 \pm 8.26$ & $30.68 \pm 15.07$ & $21.33 \pm 3.33$ \\
Upper $\delta$ OFF & $99.11 \pm 0.42$ & $70.62 \pm 8.60$ & $30.71 \pm 15.01$ & $21.29 \pm 3.75$ \\
Upper $\delta$ input-SG ON & $98.98 \pm 0.43$ & $65.25 \pm 12.40$ & $32.97 \pm 16.67$ & $22.95 \pm 2.91$ \\
Upper $\delta$ input-SG OFF & $99.01 \pm 0.47$ & $65.37 \pm 12.20$ & $33.01 \pm 16.76$ & $23.02 \pm 2.75$ \\
\bottomrule
\end{tabular}
\end{table}

The seven-seed comparison first distinguishes fitting the training-length regime from generalizing beyond it. 
All four trained configurations have mean length-40 accuracy above 98\%, yet their length-80 outcome distributions differ substantially (Table~\ref{tab:upper-seven}).
Across seeds 42--48, the local-attention recurrent baseline reaches $69.21\pm22.42$\%, compared with $70.81\pm8.26$\% for upper delta attention. 
The sample SD is 63.2\% smaller, a factor of 2.71, while the mean differs by $+1.60$ percentage points.
The baseline ranges from 33.11\% to 86.62\%; delta attention ranges from 61.52\% to 80.08\%.
Several stronger baseline seeds decrease while the two weakest improve (Figure~\ref{fig:seven-seed}).
The result is a narrower extrapolation distribution with a similar mean, rather than an improvement for every seed or evidence that training-length fitting previously failed.

\paragraph{The state-attention control shares the reduction.}
Upper state attention reaches $74.75\pm6.24$\%, with both a higher mean and a smaller SD than upper delta attention.
Thus reduced seed variation is an observed benefit of both upper-attention designs, not a distinctive property
of delta attention.
The subsequent comparisons ask how the bounded update route contributes during learning and whether it avoids the severe extrapolation degradation observed with state attention at the lower layer.

Appendix~\ref{app:seed-details} reports individual outcomes for all seven training seeds.

\paragraph{The upper delta correction is largely dispensable at evaluation.}
\label{sec:off}
Setting its scale to zero on the same seven checkpoints leaves $70.62\pm8.60$\% accuracy. The average ON--OFF difference is 0.20 percentage points and the largest absolute per-seed difference is 1.27 points.
The similar mean and reduced spread therefore largely persist without the correction at inference.
This is consistent with a contribution during learning whose benefit remains in the trained network.

\subsection{Blocking the upper input-gradient route partially reverses the variance reduction}
\label{sec:scaffold-results}
% Generated by scripts/build_artifacts.py.
\begin{table}[htbp]
\centering
\caption{Upper delta attention with and without gradients at its raw-update Q/K/V inputs, using seeds 42--48 and constant scale 0.05. ON/OFF entries are length-80 final-position accuracy (\%), mean $\pm$ sample SD over all seven seeds at LAST 40k. OFF changes only the inference-time correction scale. First 95\% and First 99\% give optimizer updates to the first ordinary length-40 validation final-position accuracy at or above each threshold, checked every 250 updates on 2,000 fixed examples; these are first crossings, not sustained attainment. The 95\% statistics use all seven seeds (42--48); the 99\% statistics use the same six reached seeds (42--47) in both rows. Seed 48 does not reach 99\% within 40k in either condition and is not imputed. $n_{95}$ and $n_{99}$ give the attained counts. Each training seed is counted once.}
\label{tab:upper-gradient}
\footnotesize
\setlength{\tabcolsep}{4pt}
\begin{tabular}{lrrrrrr}
\toprule
Training gradients & ON: $L=80$ & OFF: $L=80$ & First 95\% & $n_{95}$ & First 99\% & $n_{99}$ \\
\midrule
Full gradient & $70.81 \pm 8.26$ & $70.62 \pm 8.60$ & $17,750 \pm 2,983$ & 7/7 & $28,583 \pm 5,610$ & 6/7 \\
Input-SG & $65.25 \pm 12.40$ & $65.37 \pm 12.20$ & $17,179 \pm 3,068$ & 7/7 & $31,667 \pm 2,705$ & 6/7 \\
\bottomrule
\end{tabular}
\end{table}

We compare upper delta attention with input-SG on the same seven seeds 42--48.
Input-SG blocks all raw-update Q/K/V input gradients through correction attention while retaining its forward computation, trainable attention projections, and the main recurrent gradient path (Section~\ref{sec:scaffold-protocol}).
Parameter count and direct field match; at identical weights and inputs, the forward values are identical.

Length-80 accuracy changes from $70.81\pm8.26$\% with full gradients to $65.25\pm12.40$\% with input-SG (Table~\ref{tab:upper-gradient}).
The SD increases by a factor of 1.50 (50.2\%), alongside a 5.57-point decrease in the mean.
Full gradients yield higher accuracy in six of seven matched seeds (Table~\ref{tab:upper-gradient-seeds}).
This partially reverses the variance reduction: the input-SG SD remains below the matched baseline's 22.42 points,
a 44.7\% reduction from baseline rather than the full-gradient reduction of 63.2\%.

The difference persists after disabling the correction at evaluation:
full-gradient and input-SG checkpoints retain $70.62\pm8.60$\% and $65.37\pm12.20$\%, respectively.
This places part of the observed difference in the trained network rather than solely in the correction used for the current prediction.
Together, the interventions support a contribution from the additional input-gradient route to learned extrapolation outcomes.

The learning-speed comparison depends on the validation-accuracy threshold.
Both conditions reach 95\% length-40 validation accuracy in all seven seeds.
The first crossing occurs at $17{,}750\pm2{,}983$ updates with full gradients and $17{,}179\pm3{,}068$ with input-SG (mean $\pm$ sample SD; Table~\ref{tab:upper-gradient}). 
Input-SG reaches this threshold earlier in six of seven seeds and 571 updates earlier on average. Blocking the helper input gradients therefore does not delay mean attainment of 95\% accuracy in this comparison.

At 99\%, both conditions reach length-40 validation accuracy in the same six seeds, 42--47; seed 48 does not reach it within 40,000 updates in either condition.
Among those six seeds, the first crossing occurs at a mean of 28,583 updates with full gradients, versus 31,667 with input-SG, an average delay of 3,083 updates.
Five of the six cross later with input-SG; seed 46 crosses earlier.
Nonattainment is reported separately rather than imputed or included as a crossing time.
The SD of the 99\% first-crossing time is smaller with input-SG (2,705 versus 5,610 updates).
A more uniform threshold-crossing time therefore coexists with greater variation in extrapolation accuracy.
Learning speed, training-time variability, and extrapolation variability are distinct outcomes.
Individual crossings and BEST sensitivity appear in Appendix~\ref{app:scaffold}.

\subsection{Lower delta attention causes less extrapolation degradation than state attention}
\label{sec:lower-recovery}
% Generated by scripts/build_artifacts.py.
\begin{figure}[tb]
\centering
\paperfigure{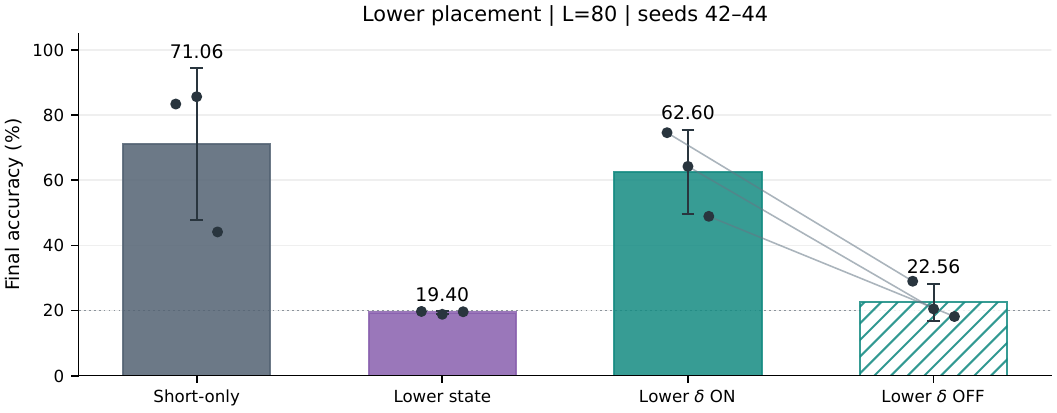}
\caption{Lower-layer extrapolation and inference-time correction dependence at $L=80$ (seeds 42--44, LAST 40k). Bars, error bars, and dots show mean accuracy, sample SD, and individual seeds. The state-read configuration is near chance (20\%); delta attention with $\alpha=0.05$ retains substantially more accuracy. Setting the trained correction scale to zero returns accuracy near chance. Lines connect ON/OFF evaluations of the same checkpoint and seed. Recurrent carry remains active in OFF. Length 80 exceeds the direct fields of all four configurations; parameter counts and constraints differ between the state- and delta-attention designs.}
\label{fig:lower-recovery}
\end{figure}

The lower-layer comparison tests the update route at a placement where a long state read severely degrades an otherwise learnable recurrent task.
For seeds 42--44, the baseline reaches $71.06\pm23.34$\% at length 80, while lower state attention reaches $19.40\pm0.48$\%, near chance.
Lower delta attention with $\alpha=0.05$ reaches $62.60\pm12.92$\% (Figure~\ref{fig:lower-recovery}).
Its mean is 43.20 percentage points above lower state attention, while remaining 8.46 points below the baseline.
The corresponding mean decrease for lower state attention is 51.66 points.
Thus the bounded update design exhibits substantially less extrapolation degradation in this placement.

\paragraph{The lower correction remains functionally necessary.}
Setting the correction scale to zero on the same lower-delta checkpoints reduces accuracy to $22.56\pm5.70$\%, a loss of 40.04 percentage points.
The recurrent carry and all other trained weights remain active.
Consequently, the smaller degradation cannot be explained by a model that simply ignores its correction: the trained network depends strongly on it at inference.
This differs from the upper-layer removal result.
The full placement and removal sweeps appear in Appendices~\ref{app:seed-details} and~\ref{app:off-details}.

Lower input-SG reaches $66.28\pm14.05$\% with the correction and $29.88\pm15.23$\% without it.
Its advantage over lower state attention and its removal sensitivity therefore persist without the additional attention-input gradient route.
The lower result cannot be attributed solely to that route.
In combination with the upper intervention, it supports examining both the forward correction and its gradient connections as placement-dependent contributors to recurrent learning.

\paragraph{Scope of the three findings.}
The findings concern length-80 generalization under the stated task and
training budget. Performance degrades substantially at longer lengths, and
checkpoint selection can change the mean ordering
(Appendix~\ref{app:checkpoint}). All-layer corrections have mixed effects
(Appendix~\ref{app:all-layer}). Scheduled upper removal and training-time
correction diagnostics are supplementary analyses in
Appendices~\ref{app:scaffold} and~\ref{app:dynamics}. The combined evidence
motivates a learning aid that can reduce observed seed variation while
causing less damage in the lower placement; it does not establish a
universal placement or a unique internal mechanism.

\section{Discussion and limitations}
\label{sec:discussion}
\paragraph{Recurrent memory requires learning useful updates.}
Our motivation is to maintain and revise information over an extended task through recurrent memory.
Latent compression methods such as AutoCompressors show that useful summaries can be learned with a language-model objective without human compression targets \citep{chevalier2023autocompressors}.
Repeatedly integrating new observations into a summary is a recurrent computation, but providing a recurrent state does not ensure that useful writes and updates are easy to learn.
The information written now may only affect a prediction much later, making temporal credit assignment a central training problem \citep{ke2018sab,kumar2026smt}.
The present study investigates an update-level learning aid for that problem, using length generalization as a diagnostic of learned recurrent behavior.

\paragraph{Attention can assist learning and offer alternative solutions.}
An attention route can connect distant computations directly while also allowing a model to solve training examples without maintaining a reusable state update.
DAT demonstrates shortcut reliance and impaired length generalization in jointly trained hybrids; SWAX shows that wider attention windows can weaken long-context memorization \citep{phan2025delayed,cabannes2025swax}. 
These findings motivate the hypothesis that learning assistance and competition with recurrent computation can coexist. 
\citet{tang2026entities} observe query-dependent aggregation rather than incremental state tracking in the studied language models.
Such query-time aggregation (QTA) may reduce pressure to learn ongoing memory updates when it suffices on the training distribution.

\paragraph{What the results establish.}
The seven-seed experiment shows a 63.2\% reduction in length-80 sample SD with a similar mean for upper delta attention.
This concerns variation in extrapolation outcomes among models with high training-length accuracy.
Upper state attention has a higher mean and smaller SD, so attention-based assistance is a shared observation rather than a benefit unique to delta attention.
The research question is how to obtain that assistance while limiting degradation of recurrent generalization.
In the matched seven-seed comparison, input-SG lowers mean length-80 accuracy from 70.81\% to 65.25\% and increases sample SD from 8.26 to 12.40 points.
It partially reverses the upper delta reduction, including after correction removal, which supports a role for the additional gradient connection during learning.
Improved credit assignment is a plausible explanation, but the experiments do not establish it as the cause of the baseline's seed spread.

\paragraph{The lower comparison diagnoses the coupling to recurrence.}
Lower state attention performs near chance on a task the baseline can learn and extrapolate, whereas the bounded update design causes much less extrapolation degradation.
This comparison remains informative even if an upper-only configuration is preferred in later applications:
it tests the  helper where the state-read design is most damaging in our experiments.
The lower delta correction is actively used, as its removal eliminates most of the length-80 performance.
Its advantage over lower state attention also survives input-gradient blocking.
Thus a purely backward-shortcut explanation is insufficient; the forward correction, its bounded interface, and learning through the remaining paths can contribute.
State-based attention can also influence future recurrence in attentive RNNs \citep{bahdanau2015align,ke2018sab}, and our state-read control can do so through subsequent layers.
The relevant distinction is the implemented coupling to the update stream, not a universal claim that state attention has no consequences for future states.

\paragraph{Bounding a correction while retaining temporal connections.}
In \model, attention accesses uncorrected proposals, and its output enters
the increment that is subsequently accumulated. The correction therefore
affects later states within the same layer. Phase coordinates make its
per-update bound directly interpretable as an angular displacement. This
provides a simple way to restrict the correction available at each step
while preserving content-dependent paths among proposals. It may encourage
useful uncorrected writes, but the bound does not guarantee that the helper
is secondary: small corrections can accumulate, change higher-layer
proposals, and be amplified by later projections. Upper OFF retention is
consistent with a learning scaffold, yet it also occurs with input-SG, and
fixed-scale diagnostics do not show correction amplitudes vanishing.
Moreover, the phase carry already has an identity derivative away from wrap
boundaries. The proposed route changes available credit-assignment paths;
it is not needed to create the first non-vanishing carry gradient.

\paragraph{Toward task-oriented recurrent memory.}
A prospective design can retain short local attention and use a limited
number of bounded update helpers while evaluating whether long state reads
provide additional task-level benefits. If sustained task performance can
be learned through this route, internal exact retrieval need not dictate
the architecture; external retrieval could supply historical details when
needed. This is a design objective, not a tested retrieval system. The
upper-only and mixed all-layer results motivate selective placement rather
than repeating the helper at every depth. Lower corrections change the
representations from which higher layers form updates, so multiple helpers
may interact even when individually bounded. That explanation remains
hypothetical. The next question is whether useful recurrent memory can be
learned under a language-model objective; the present experiment does not
demonstrate semantic compaction or sustained natural-language task quality.

\section{Conclusion}
Taken together, our results suggest that strengthening temporal skip
connections in memory-write computations is a promising direction for
improving credit assignment in recurrent models. Bounded delta attention
connects raw update proposals and constrains the correction before it
enters recurrence. On prefix-supervised modulo-5, upper delta attention
reduces seed variation while retaining a baseline-like mean, and this
behavior largely persists after removing the correction at inference.
Blocking gradients through the attention inputs partially reverses this
effect, supporting the value of feedback paths that reach the computations
producing memory writes.

The lower-layer comparison provides complementary evidence: directing
bounded attention to updates causes substantially less extrapolation
degradation than attending to accumulated state representations. The lower
model still relies on its correction at inference, indicating that useful
forward corrections and training-time feedback need to be considered
together. These findings motivate attention-based shortcuts within the
memory-writing process as an alternative to long state reads for assisting
recurrent learning. Extending this approach to recurrent memory trained
with a language-model objective is a natural next step.

\bibliographystyle{unsrtnat}
\begingroup
\raggedright
\bibliography{references}
\endgroup
\clearpage
\appendix
\section{Seed-level results and supplementary placement comparisons}
\label{app:seed-details}
Tables~\ref{tab:seed-upper-last} and~\ref{tab:seed-upper-best}
report the upper-comparison results for all seven seeds 42--48.
The supplementary placement, scale, and scaffold comparisons use the same
seeds 42--44 for every condition, including upper-layer settings. The baseline
and upper rows reuse the corresponding subset of the seven-seed results;
they are not additional independent runs. These are paired
training-seed comparisons;
the common seed controls initialization and training streams as described
in the experimental protocol. Pairing does not make different architectures
follow the same optimization trajectory.
% Generated by scripts/build_artifacts.py.
\begin{figure}[tb]
\centering
\paperfigure{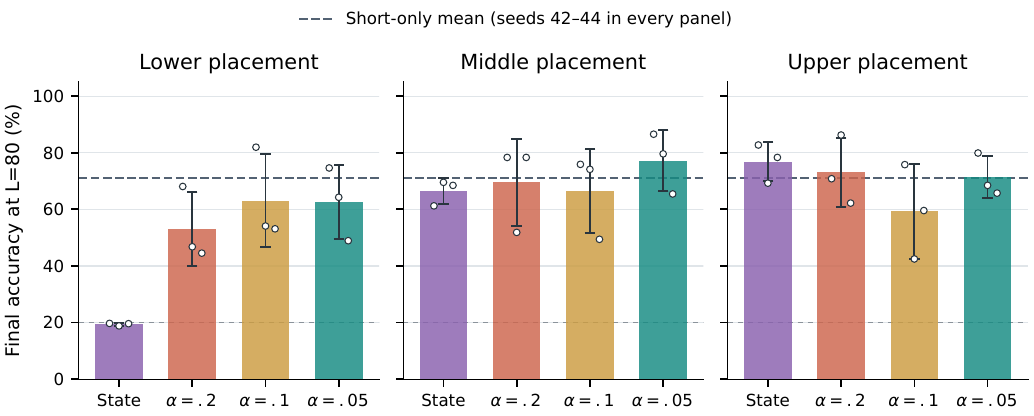}
\caption{Attention source, placement, and correction scale at evaluation length $L=80$. State denotes a read window of 40 in one layer; each $\delta$ condition instead keeps read and write windows at 5 and adds a window-40 bounded correction to raw phase increments in the indicated layer. Bars show means, error bars sample standard deviations, and points seeds 42--44 at LAST (40k updates). All placements use the same three seeds, including the upper state and scale-0.05 subset of Table~\ref{tab:upper-seven}. The short-only reference mean is reused across panels. $L=80$ exceeds the short-only, state, and single-$\delta$ direct fields (31, 66, and 71, respectively). Comparisons do not match parameter count, source representation, or correction constraint; the figure therefore does not isolate attention source as the sole causal factor. The dotted gray line is 20\% chance accuracy.}
\label{fig:placement-scale}
\end{figure}

% Generated by scripts/build_artifacts.py.
\begin{table}[htbp]
\centering
\caption{Supplementary lower/middle/upper placement comparisons at LAST (40k updates). Values are final-position accuracy in percent, reported as mean $\pm$ sample standard deviation across the same seeds 42--44 in every row. Baseline and upper rows reuse the corresponding subset of Table~\ref{tab:upper-seven}. The state and $\delta$ models are not parameter-matched; Table~\ref{tab:controls} gives their distinct interventions and direct fields.}
\label{tab:primary-placement}
\small
\setlength{\tabcolsep}{4pt}
\begin{tabular}{llrrrr}
\toprule
Condition & $\alpha$ & $L=40$ & $L=80$ & $L=160$ & $L=320$ \\
\midrule
Short-only & --- & $99.09 \pm 0.84$ & $71.06 \pm 23.34$ & $33.92 \pm 21.47$ & $25.13 \pm 8.02$ \\
Lower state & --- & $98.37 \pm 0.06$ & $19.40 \pm 0.48$ & $20.21 \pm 1.81$ & $19.63 \pm 2.11$ \\
Lower $\delta$ & 0.05 & $98.31 \pm 1.24$ & $62.60 \pm 12.92$ & $22.72 \pm 5.83$ & $20.15 \pm 1.75$ \\
Middle state & --- & $99.02 \pm 0.00$ & $66.41 \pm 4.51$ & $20.28 \pm 0.73$ & $20.25 \pm 0.66$ \\
Middle $\delta$ & 0.05 & $99.12 \pm 0.35$ & $77.18 \pm 10.75$ & $34.18 \pm 19.82$ & $23.05 \pm 7.20$ \\
Upper state & --- & $99.45 \pm 0.72$ & $76.76 \pm 6.87$ & $30.76 \pm 19.29$ & $23.93 \pm 7.53$ \\
Upper $\delta$ & 0.05 & $99.35 \pm 0.50$ & $71.35 \pm 7.51$ & $25.62 \pm 12.19$ & $19.37 \pm 1.17$ \\
\bottomrule
\end{tabular}
\end{table}

Figure~\ref{fig:placement-scale} and Table~\ref{tab:primary-placement} give the
supplementary placement and scale comparisons. With $\alpha=0.05$,
middle delta attention reaches $77.18\pm10.75$\% at length 80.
The scale sweep is not monotone: smaller correction scales do not always
improve extrapolation. Upper scale 0.05 was selected for the seed extension
because it combined a baseline-like mean with a smaller observed spread.

% Generated by scripts/build_artifacts.py.
\begin{table}[htbp]
\centering
\caption{Individual-seed upper-layer final-position accuracies (\%), LAST checkpoints. BEST denotes the minimum ordinary length-40 validation cross-entropy within the completed 40k run.}
\label{tab:seed-upper-last}
\footnotesize
\setlength{\tabcolsep}{4pt}
\begin{tabular}{rlrrrr}
\toprule
Seed & Condition & $L=40$ & $L=80$ & $L=160$ & $L=320$ \\
\midrule
42 & Short-only & 98.93 & 83.40 & 58.59 & 34.28 \\
42 & Upper state & 98.63 & 82.71 & 53.03 & 32.62 \\
42 & Upper $\delta$ ON & 99.22 & 79.88 & 39.65 & 20.51 \\
42 & Upper $\delta$ OFF & 99.32 & 79.88 & 39.55 & 20.31 \\
42 & Upper $\delta$ input-SG ON & 99.22 & 82.03 & 54.49 & 25.39 \\
42 & Upper $\delta$ input-SG OFF & 99.22 & 81.84 & 54.49 & 25.39 \\
43 & Short-only & 100.00 & 85.64 & 23.63 & 19.34 \\
43 & Upper state & 99.71 & 69.24 & 19.04 & 19.92 \\
43 & Upper $\delta$ ON & 99.90 & 68.46 & 19.53 & 18.16 \\
43 & Upper $\delta$ OFF & 99.71 & 68.26 & 19.82 & 17.09 \\
43 & Upper $\delta$ input-SG ON & 98.24 & 66.11 & 22.95 & 20.41 \\
43 & Upper $\delta$ input-SG OFF & 98.24 & 66.41 & 22.56 & 20.61 \\
44 & Short-only & 98.34 & 44.14 & 19.53 & 21.78 \\
44 & Upper state & 100.00 & 78.32 & 20.21 & 19.24 \\
44 & Upper $\delta$ ON & 98.93 & 65.72 & 17.68 & 19.43 \\
44 & Upper $\delta$ OFF & 98.83 & 65.14 & 17.58 & 20.12 \\
44 & Upper $\delta$ input-SG ON & 99.41 & 57.91 & 19.53 & 22.27 \\
44 & Upper $\delta$ input-SG OFF & 99.41 & 58.30 & 20.12 & 21.97 \\
45 & Short-only & 99.02 & 86.33 & 53.52 & 25.68 \\
45 & Upper state & 99.41 & 82.23 & 48.44 & 30.37 \\
45 & Upper $\delta$ ON & 99.32 & 77.83 & 41.89 & 21.88 \\
45 & Upper $\delta$ OFF & 99.32 & 78.12 & 42.09 & 22.36 \\
45 & Upper $\delta$ input-SG ON & 99.32 & 67.87 & 39.45 & 22.07 \\
45 & Upper $\delta$ input-SG OFF & 99.32 & 68.26 & 39.45 & 21.88 \\
46 & Short-only & 98.14 & 33.11 & 18.36 & 21.29 \\
46 & Upper state & 99.71 & 68.16 & 18.85 & 19.63 \\
46 & Upper $\delta$ ON & 98.83 & 62.21 & 22.46 & 19.53 \\
46 & Upper $\delta$ OFF & 98.73 & 60.94 & 22.66 & 19.43 \\
46 & Upper $\delta$ input-SG ON & 99.02 & 50.59 & 19.92 & 21.29 \\
46 & Upper $\delta$ input-SG OFF & 99.22 & 50.59 & 20.02 & 21.29 \\
47 & Short-only & 98.44 & 86.62 & 33.01 & 17.68 \\
47 & Upper state & 96.48 & 72.66 & 26.86 & 19.92 \\
47 & Upper $\delta$ ON & 99.32 & 80.08 & 55.76 & 28.32 \\
47 & Upper $\delta$ OFF & 99.32 & 80.27 & 55.57 & 29.00 \\
47 & Upper $\delta$ input-SG ON & 99.12 & 79.59 & 55.57 & 28.42 \\
47 & Upper $\delta$ input-SG OFF & 99.22 & 79.20 & 55.96 & 28.22 \\
48 & Short-only & 98.44 & 65.23 & 26.37 & 20.41 \\
48 & Upper state & 99.71 & 69.92 & 22.46 & 19.53 \\
48 & Upper $\delta$ ON & 98.63 & 61.52 & 17.77 & 21.48 \\
48 & Upper $\delta$ OFF & 98.54 & 61.72 & 17.68 & 20.70 \\
48 & Upper $\delta$ input-SG ON & 98.54 & 52.64 & 18.85 & 20.80 \\
48 & Upper $\delta$ input-SG OFF & 98.44 & 53.03 & 18.46 & 21.78 \\
\bottomrule
\end{tabular}
\end{table}

% Generated by scripts/build_artifacts.py.
\begin{table}[htbp]
\centering
\caption{Individual-seed upper-layer final-position accuracies (\%), BEST checkpoints. BEST denotes the minimum ordinary length-40 validation cross-entropy within the completed 40k run.}
\label{tab:seed-upper-best}
\footnotesize
\setlength{\tabcolsep}{4pt}
\begin{tabular}{rlrrrr}
\toprule
Seed & Condition & $L=40$ & $L=80$ & $L=160$ & $L=320$ \\
\midrule
42 & Short-only & 99.32 & 86.82 & 59.57 & 34.96 \\
42 & Upper state & 98.93 & 83.20 & 55.18 & 32.52 \\
42 & Upper $\delta$ ON & 99.32 & 86.13 & 50.78 & 28.61 \\
42 & Upper $\delta$ OFF & 99.32 & 86.04 & 51.07 & 28.61 \\
42 & Upper $\delta$ input-SG ON & 99.12 & 79.00 & 57.13 & 33.50 \\
42 & Upper $\delta$ input-SG OFF & 99.22 & 79.39 & 56.84 & 33.79 \\
43 & Short-only & 100.00 & 87.01 & 25.39 & 18.55 \\
43 & Upper state & 99.61 & 71.58 & 21.88 & 20.21 \\
43 & Upper $\delta$ ON & 100.00 & 76.27 & 21.68 & 18.07 \\
43 & Upper $\delta$ OFF & 100.00 & 75.39 & 21.97 & 17.87 \\
43 & Upper $\delta$ input-SG ON & 99.02 & 55.57 & 23.14 & 19.73 \\
43 & Upper $\delta$ input-SG OFF & 98.93 & 55.57 & 23.24 & 20.21 \\
44 & Short-only & 99.51 & 57.71 & 18.65 & 21.09 \\
44 & Upper state & 99.90 & 78.03 & 20.21 & 18.26 \\
44 & Upper $\delta$ ON & 99.32 & 58.79 & 18.95 & 18.36 \\
44 & Upper $\delta$ OFF & 99.41 & 58.50 & 18.36 & 18.65 \\
44 & Upper $\delta$ input-SG ON & 99.41 & 45.70 & 20.02 & 20.21 \\
44 & Upper $\delta$ input-SG OFF & 99.41 & 46.97 & 20.70 & 20.31 \\
45 & Short-only & 99.02 & 88.28 & 54.79 & 26.86 \\
45 & Upper state & 99.22 & 85.55 & 52.93 & 32.52 \\
45 & Upper $\delta$ ON & 99.32 & 78.32 & 43.95 & 21.97 \\
45 & Upper $\delta$ OFF & 99.32 & 78.42 & 44.14 & 22.75 \\
45 & Upper $\delta$ input-SG ON & 99.12 & 67.68 & 39.26 & 21.78 \\
45 & Upper $\delta$ input-SG OFF & 99.12 & 67.58 & 38.87 & 21.09 \\
46 & Short-only & 99.22 & 36.82 & 18.16 & 21.29 \\
46 & Upper state & 99.32 & 69.92 & 18.65 & 18.95 \\
46 & Upper $\delta$ ON & 98.83 & 62.21 & 22.46 & 19.53 \\
46 & Upper $\delta$ OFF & 98.73 & 60.94 & 22.66 & 19.43 \\
46 & Upper $\delta$ input-SG ON & 99.02 & 43.65 & 20.70 & 21.29 \\
46 & Upper $\delta$ input-SG OFF & 99.02 & 44.63 & 20.31 & 21.29 \\
47 & Short-only & 99.41 & 89.94 & 33.59 & 19.73 \\
47 & Upper state & 99.02 & 85.55 & 42.48 & 19.73 \\
47 & Upper $\delta$ ON & 99.41 & 79.98 & 54.00 & 25.10 \\
47 & Upper $\delta$ OFF & 99.41 & 79.88 & 53.42 & 25.20 \\
47 & Upper $\delta$ input-SG ON & 99.12 & 75.68 & 52.44 & 28.22 \\
47 & Upper $\delta$ input-SG OFF & 99.12 & 75.39 & 52.44 & 28.12 \\
48 & Short-only & 98.83 & 77.83 & 21.97 & 20.41 \\
48 & Upper state & 99.71 & 68.75 & 21.88 & 17.48 \\
48 & Upper $\delta$ ON & 98.73 & 60.64 & 16.02 & 21.09 \\
48 & Upper $\delta$ OFF & 98.63 & 60.64 & 16.50 & 20.41 \\
48 & Upper $\delta$ input-SG ON & 98.63 & 64.84 & 17.38 & 18.95 \\
48 & Upper $\delta$ input-SG OFF & 98.63 & 64.84 & 17.58 & 18.95 \\
\bottomrule
\end{tabular}
\end{table}

% Generated by scripts/build_artifacts.py.
\begin{table}[htbp]
\centering
\caption{Individual-seed lower-layer comparisons at LAST 40k. Values are final-position accuracy (\%).}
\label{tab:seed-lower}
\footnotesize
\setlength{\tabcolsep}{4pt}
\begin{tabular}{llrrrr}
\toprule
Condition & Seed & $L=40$ & $L=80$ & $L=160$ & $L=320$ \\
\midrule
State & 42 & 98.34 & 19.73 & 18.36 & 20.21 \\
State & 43 & 98.34 & 18.85 & 20.31 & 17.29 \\
State & 44 & 98.44 & 19.63 & 21.97 & 21.39 \\
$\delta$, $\alpha=0.2$ & 42 & 99.12 & 68.07 & 22.66 & 19.34 \\
$\delta$, $\alpha=0.2$ & 43 & 99.41 & 46.78 & 18.75 & 23.05 \\
$\delta$, $\alpha=0.2$ & 44 & 98.83 & 44.53 & 19.43 & 19.34 \\
$\delta$, $\alpha=0.1$ & 42 & 99.41 & 81.93 & 31.93 & 19.53 \\
$\delta$, $\alpha=0.1$ & 43 & 96.39 & 54.10 & 20.61 & 22.46 \\
$\delta$, $\alpha=0.1$ & 44 & 99.32 & 53.12 & 21.48 & 19.92 \\
$\delta$, $\alpha=0.05$ & 42 & 98.93 & 74.61 & 29.00 & 18.65 \\
$\delta$, $\alpha=0.05$ & 43 & 96.88 & 64.26 & 17.48 & 19.73 \\
$\delta$, $\alpha=0.05$ & 44 & 99.12 & 48.93 & 21.68 & 22.07 \\
\bottomrule
\end{tabular}
\end{table}

% Generated by scripts/build_artifacts.py.
\begin{table}[htbp]
\centering
\caption{Individual-seed middle-layer comparisons at LAST 40k. Values are final-position accuracy (\%).}
\label{tab:seed-middle}
\footnotesize
\setlength{\tabcolsep}{4pt}
\begin{tabular}{llrrrr}
\toprule
Condition & Seed & $L=40$ & $L=80$ & $L=160$ & $L=320$ \\
\midrule
State & 42 & 99.02 & 61.23 & 20.31 & 19.73 \\
State & 43 & 99.02 & 69.53 & 19.53 & 20.02 \\
State & 44 & 99.02 & 68.46 & 21.00 & 21.00 \\
$\delta$, $\alpha=0.2$ & 42 & 99.51 & 78.32 & 40.43 & 19.43 \\
$\delta$, $\alpha=0.2$ & 43 & 95.12 & 51.86 & 16.89 & 21.19 \\
$\delta$, $\alpha=0.2$ & 44 & 99.41 & 78.32 & 17.97 & 20.80 \\
$\delta$, $\alpha=0.1$ & 42 & 99.22 & 75.88 & 39.84 & 27.44 \\
$\delta$, $\alpha=0.1$ & 43 & 99.51 & 74.12 & 20.90 & 19.73 \\
$\delta$, $\alpha=0.1$ & 44 & 98.44 & 49.41 & 21.19 & 20.02 \\
$\delta$, $\alpha=0.05$ & 42 & 99.51 & 86.52 & 57.03 & 31.35 \\
$\delta$, $\alpha=0.05$ & 43 & 99.02 & 79.59 & 23.93 & 18.46 \\
$\delta$, $\alpha=0.05$ & 44 & 98.83 & 65.43 & 21.58 & 19.34 \\
\bottomrule
\end{tabular}
\end{table}

% Generated by scripts/build_artifacts.py.
\begin{table}[htbp]
\centering
\caption{Individual-seed upper-layer comparisons at LAST 40k. Values are final-position accuracy (\%). Seeds 42--44 are shown for every condition to match the lower/middle comparisons; upper state and scale 0.05 reuse a subset of Table~\ref{tab:upper-seven}.}
\label{tab:seed-upper}
\footnotesize
\setlength{\tabcolsep}{4pt}
\begin{tabular}{llrrrr}
\toprule
Condition & Seed & $L=40$ & $L=80$ & $L=160$ & $L=320$ \\
\midrule
State & 42 & 98.63 & 82.71 & 53.03 & 32.62 \\
State & 43 & 99.71 & 69.24 & 19.04 & 19.92 \\
State & 44 & 100.00 & 78.32 & 20.21 & 19.24 \\
$\delta$, $\alpha=0.2$ & 42 & 99.41 & 70.80 & 48.93 & 24.41 \\
$\delta$, $\alpha=0.2$ & 43 & 100.00 & 86.23 & 21.09 & 17.58 \\
$\delta$, $\alpha=0.2$ & 44 & 99.51 & 62.21 & 20.61 & 21.19 \\
$\delta$, $\alpha=0.1$ & 42 & 99.41 & 75.78 & 54.79 & 28.42 \\
$\delta$, $\alpha=0.1$ & 43 & 97.66 & 42.48 & 19.63 & 19.43 \\
$\delta$, $\alpha=0.1$ & 44 & 99.22 & 59.57 & 18.26 & 20.02 \\
$\delta$, $\alpha=0.05$ & 42 & 99.22 & 79.88 & 39.65 & 20.51 \\
$\delta$, $\alpha=0.05$ & 43 & 99.90 & 68.46 & 19.53 & 18.16 \\
$\delta$, $\alpha=0.05$ & 44 & 98.93 & 65.72 & 17.68 & 19.43 \\
\bottomrule
\end{tabular}
\end{table}

\section{Length and checkpoint sensitivity}
\label{app:checkpoint}
% Generated by scripts/build_artifacts.py.
\begin{figure}[tb]
\centering
\paperfigure{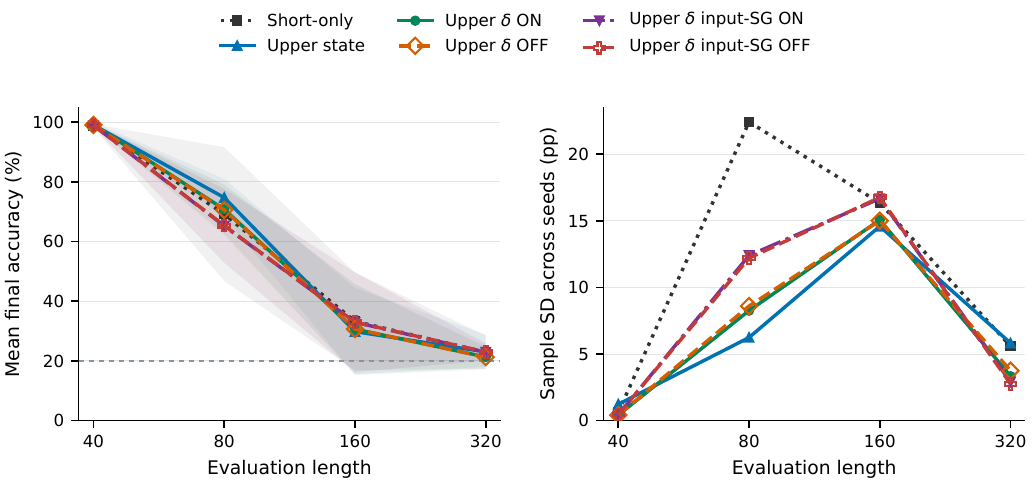}
\caption{Seven-seed upper-layer results across evaluation lengths at LAST. Left: mean final-position accuracy with one sample standard deviation shaded and clipped to [0,100]. Right: the same standard deviations in percentage points. The training-length boundary is 40; all displayed longer lengths exceed the relevant direct fields. A small spread near chance is not evidence of successful recurrence. Distinct colors, line styles, and markers identify conditions; OFF uses open markers. At $L=80$, upper state, upper $\delta$, and upper input-SG have smaller sample SDs than the short-only reference. The full-gradient advantage over input-SG at length 80 does not extend uniformly to longer lengths.}
\label{fig:upper-length-curves}
\end{figure}

Figure~\ref{fig:upper-length-curves} shows substantial degradation beyond
length 80. At length 160, the seven-seed baseline, upper delta, and upper
state means are 33.29\%, 30.68\%, and 29.84\%; at length 320 they are
22.92\%, 21.33\%, and 23.03\%. A small SD near chance should not be read
as reliable recurrent generalization.
Upper input-SG reaches $32.97\pm16.67$\% at length 160 and
$22.95\pm2.91$\% at length 320. The full-gradient mean advantage at length
80 therefore does not extend uniformly to longer evaluation sequences.

Checkpoint selection also changes the mean comparison. Using each run's
minimum in-distribution validation-cross-entropy checkpoint, the length-80
means are 74.92\% for the baseline and 71.76\% for upper delta attention,
with SDs of 20.20 and 10.96 points. The reduced spread persists, but the
baseline-like mean at the final checkpoint does not imply a mean advantage
under every checkpoint rule.
Upper input-SG reaches $61.73\pm13.90$\% at the validation-selected BEST
checkpoints, compared with $71.76\pm10.96$\% for full gradients; the
length-80 mean decrease and SD increase therefore also appear under this
checkpoint rule in the seven-seed comparison.

% Generated by scripts/build_artifacts.py.
\begin{table}[htbp]
\centering
\caption{Seven-seed checkpoint-selection sensitivity. LAST is the primary fixed-budget endpoint; BEST is selected by ordinary length-40 validation cross-entropy, not by extrapolation performance. Entries are accuracy (\%), mean $\pm$ sample standard deviation. OFF is evaluated on the corresponding ON checkpoint.}
\label{tab:checkpoint-sensitivity}
\small
\setlength{\tabcolsep}{4pt}
\begin{tabular}{lrrrr}
\toprule
Condition & LAST: $L=80$ & BEST: $L=80$ & LAST: $L=160$ & BEST: $L=160$ \\
\midrule
Short-only & $69.21 \pm 22.42$ & $74.92 \pm 20.20$ & $33.29 \pm 16.34$ & $33.16 \pm 17.25$ \\
Upper state & $74.75 \pm 6.24$ & $77.51 \pm 7.43$ & $29.84 \pm 14.59$ & $33.31 \pm 16.30$ \\
Upper $\delta$ ON & $70.81 \pm 8.26$ & $71.76 \pm 10.96$ & $30.68 \pm 15.07$ & $32.55 \pm 16.33$ \\
Upper $\delta$ OFF & $70.62 \pm 8.60$ & $71.40 \pm 11.13$ & $30.71 \pm 15.01$ & $32.59 \pm 16.24$ \\
Upper $\delta$ input-SG ON & $65.25 \pm 12.40$ & $61.73 \pm 13.90$ & $32.97 \pm 16.67$ & $32.87 \pm 16.63$ \\
Upper $\delta$ input-SG OFF & $65.37 \pm 12.20$ & $62.05 \pm 13.48$ & $33.01 \pm 16.76$ & $32.85 \pm 16.46$ \\
\bottomrule
\end{tabular}
\end{table}

BEST uses only in-distribution validation cross-entropy. No extrapolation
metric selects a checkpoint, and runs may reach BEST at different steps.

\section{Correction removal across placements and scales}
\label{app:off-details}
% Generated by scripts/build_artifacts.py.
\begin{figure}[tb]
\centering
\paperfigure{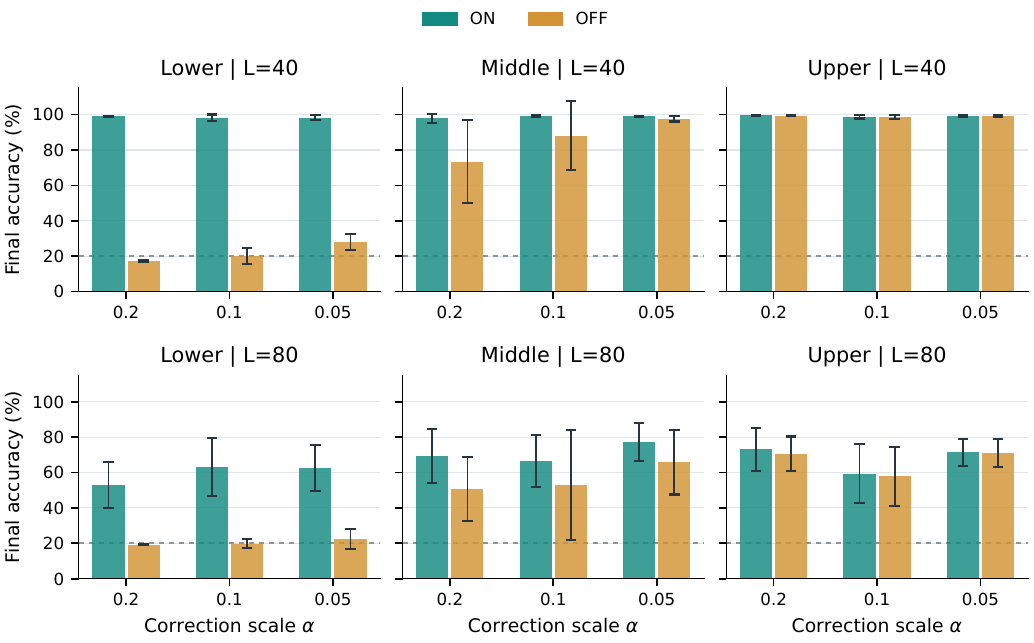}
\caption{Inference-time correction removal across placement and scale, using the same LAST weights before and after removal. Bars show means over the same seeds 42--44 in every panel; error bars show sample standard deviations. Upper scale 0.05 uses the corresponding three-seed subset of Table~\ref{tab:upper-seven}. OFF sets $\alpha=0$ but retains recurrent phase inheritance and every local read/write operation. The effect depends strongly on placement: lower-layer models can depend on the correction even at $L=40$, whereas upper-layer ON and OFF results are close. This intervention diagnoses dependence on the learned correction at inference; it does not establish how training would behave without that correction.}
\label{fig:off-placement}
\end{figure}

% Generated by scripts/build_artifacts.py.
\begin{table}[htbp]
\centering
\caption{Inference-time correction removal across placement and scale at LAST. Mean accuracy (\%) $\pm$ sample standard deviation over seeds 42--44. OFF sets the correction scale to zero while retaining the trained parameters, recurrent inheritance, and local read/write paths. Upper scale 0.05 uses the same three seeds as the other settings, a subset of Table~\ref{tab:upper-seven}.}
\label{tab:off-placement}
\small
\setlength{\tabcolsep}{4pt}
\begin{tabular}{llrrrr}
\toprule
Placement & $\alpha$ & ON: $L=40$ & OFF: $L=40$ & ON: $L=80$ & OFF: $L=80$ \\
\midrule
Lower & 0.2 & $99.12 \pm 0.29$ & $17.15 \pm 0.59$ & $53.12 \pm 12.99$ & $19.17 \pm 0.15$ \\
Lower & 0.1 & $98.37 \pm 1.72$ & $20.12 \pm 4.75$ & $63.05 \pm 16.36$ & $19.82 \pm 2.71$ \\
Lower & 0.05 & $98.31 \pm 1.24$ & $28.06 \pm 4.47$ & $62.60 \pm 12.92$ & $22.56 \pm 5.70$ \\
Middle & 0.2 & $98.01 \pm 2.51$ & $73.44 \pm 23.55$ & $69.50 \pm 15.28$ & $50.68 \pm 18.15$ \\
Middle & 0.1 & $99.06 \pm 0.56$ & $88.05 \pm 19.51$ & $66.47 \pm 14.80$ & $52.99 \pm 31.16$ \\
Middle & 0.05 & $99.12 \pm 0.35$ & $97.69 \pm 1.58$ & $77.18 \pm 10.75$ & $65.76 \pm 18.14$ \\
Upper & 0.2 & $99.64 \pm 0.31$ & $99.32 \pm 0.26$ & $73.08 \pm 12.17$ & $70.74 \pm 9.62$ \\
Upper & 0.1 & $98.76 \pm 0.96$ & $98.63 \pm 1.19$ & $59.28 \pm 16.65$ & $57.81 \pm 16.55$ \\
Upper & 0.05 & $99.35 \pm 0.50$ & $99.28 \pm 0.44$ & $71.35 \pm 7.51$ & $71.09 \pm 7.77$ \\
\bottomrule
\end{tabular}
\end{table}

OFF sets the trained correction scale to zero and keeps the recurrent carry
and all other trained weights active. This intervention uses the same
evaluation examples as ON. It is not a separately trained model without
the auxiliary route. The contrast between the larger lower-layer losses
and small upper-layer changes shows that removal behavior depends on
placement and scale. It does not prove that every individual prediction
or hidden trajectory is unchanged when aggregate accuracy is retained.

\section{All-layer corrections}
\label{app:all-layer}
% Generated by scripts/build_artifacts.py.
\begin{figure}[tb]
\centering
\paperfigure{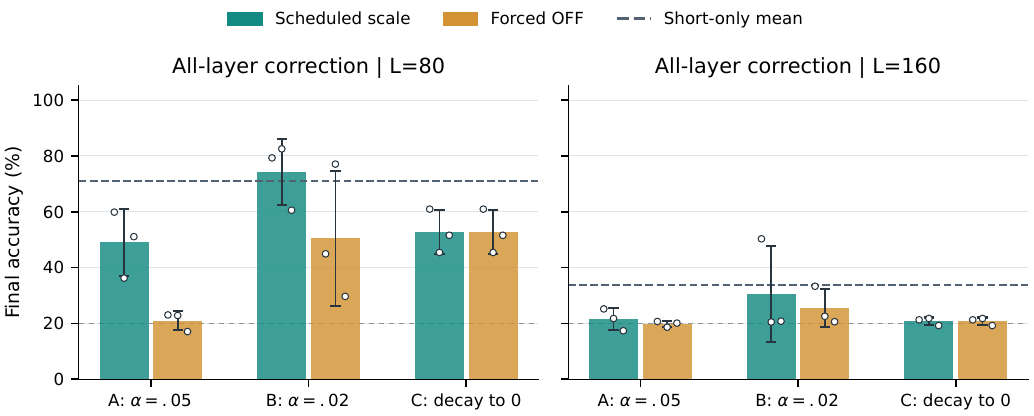}
\caption{Supplementary all-layer correction experiments (seeds 42--44, LAST 40k). A and B keep scales 0.05 and 0.02 constant in all three layers. C starts at 0.05, decays linearly from step 24k to 36k, and is already zero at LAST; its scheduled and forced-OFF evaluations therefore coincide. With nonzero correction, the direct field is 151: $L=80$ is within that field, while $L=160$ is beyond it. With correction disabled the direct field is 31. Bars, error bars, and points show means, sample standard deviations, and seeds. B reaches 74.12\% at $L=80$, within its direct field, and 30.53\% at $L=160$; the corresponding short-only means are 71.06\% and 33.92\%. Increasing the number of correction layers does not uniformly improve extrapolation.}
\label{fig:all-layer}
\end{figure}

% Generated by scripts/build_artifacts.py.
\begin{table}[htbp]
\centering
\caption{Supplementary all-layer experiments at LAST 40k, seeds 42--44. Every layer has a window-40 correction branch. C decays linearly between steps 24k and 36k and has zero correction at LAST. Its scheduled and OFF rows are therefore identical. Entries are mean accuracy (\%) $\pm$ sample standard deviation.}
\label{tab:all-layer}
\footnotesize
\setlength{\tabcolsep}{4pt}
\begin{tabular}{llrrrr}
\toprule
Schedule & Evaluation & $L=40$ & $L=80$ & $L=160$ & $L=320$ \\
\midrule
A: Constant 0.05 & Scheduled & $98.50 \pm 0.98$ & $49.06 \pm 11.94$ & $21.45 \pm 3.92$ & $19.21 \pm 0.41$ \\
A: Constant 0.05 & OFF & $25.72 \pm 0.89$ & $21.00 \pm 3.38$ & $19.82 \pm 1.06$ & $19.27 \pm 1.27$ \\
B: Constant 0.02 & Scheduled & $99.32 \pm 0.61$ & $74.12 \pm 11.87$ & $30.53 \pm 17.11$ & $21.78 \pm 1.57$ \\
B: Constant 0.02 & OFF & $66.47 \pm 21.14$ & $50.55 \pm 24.18$ & $25.49 \pm 6.84$ & $21.06 \pm 1.02$ \\
C: 0.05, decay to 0 & Scheduled & $98.40 \pm 0.31$ & $52.64 \pm 7.82$ & $20.77 \pm 1.35$ & $18.42 \pm 1.66$ \\
C: 0.05, decay to 0 & OFF & $98.40 \pm 0.31$ & $52.64 \pm 7.82$ & $20.77 \pm 1.35$ & $18.42 \pm 1.66$ \\
\bottomrule
\end{tabular}
\end{table}

All three layers receive a window-40 delta-attention module. The constant
schedules use scale 0.05 (A) or 0.02 (B). The decaying schedule (C) holds
0.05 through step 24,000, decreases it linearly to zero at step 36,000, and
trains the remaining 4,000 steps with zero correction. This scheduled
removal differs from evaluation-only OFF on the constant-scale checkpoints.

At length 80, constant scale 0.05 and the decaying schedule yield
$49.06\pm11.94$\% and $52.64\pm7.82$\%. Constant scale 0.02 reaches
$74.12\pm11.87$\%. On the same seeds 42--44, upper-only scale 0.05 reaches
$71.35\pm7.51$\% (Table~\ref{tab:primary-placement}); these supplementary
all-layer experiments do not establish a consistent improvement in both
accuracy and spread.

The direct field spans 151 positions with active corrections and returns
to 31 with zero correction. Thus length-80 accuracy alone cannot establish
use of recurrent carry in the active all-layer designs. The schedule
changes both correction strength and direct access; these comparisons do
not isolate attention count or exhaust possible schedules.

\section{Scaffold ablations: seeds, sensitivity, and measurement checks}
\label{app:scaffold}
% Generated by scripts/build_artifacts.py.
\begin{table}[htbp]
\centering
\caption{Matched seven-seed upper input-gradient intervention. Accuracies (\%) are ordinary length-80 final-position scores at LAST 40k; ON/OFF share each trained checkpoint. The last four columns give optimizer updates to the first ordinary length-40 validation final-position accuracy of at least 95\% or 99\%, checked every 250 updates on 2,000 fixed examples. All seven seeds reach 95\% in both conditions; seed 48 does not reach 99\% in either condition. These are first crossings, not sustained attainment. A Not reached entry denotes nonattainment within the common budget. Input-SG blocks the helper input gradients while retaining trainable attention projections and the main recurrent route.}
\label{tab:upper-gradient-seeds}
\footnotesize
\setlength{\tabcolsep}{4pt}
\begin{tabular}{rrrrrrrrr}
\toprule
Seed & $\delta$ ON & SG ON & $\delta$ OFF & SG OFF & $\delta$: 95\% & SG: 95\% & $\delta$: 99\% & SG: 99\% \\
\midrule
42 & 79.88 & 82.03 & 79.88 & 81.84 & 17,750 & 15,500 & 29,500 & 33,000 \\
43 & 68.46 & 66.11 & 68.26 & 66.41 & 13,250 & 21,250 & 21,250 & 32,750 \\
44 & 65.72 & 57.91 & 65.14 & 58.30 & 19,250 & 18,000 & 28,500 & 33,250 \\
45 & 77.83 & 67.87 & 78.12 & 68.26 & 14,750 & 12,000 & 24,500 & 26,250 \\
46 & 62.21 & 50.59 & 60.94 & 50.59 & 20,500 & 18,750 & 37,750 & 33,000 \\
47 & 80.08 & 79.59 & 80.27 & 79.20 & 17,250 & 15,500 & 30,000 & 31,750 \\
48 & 61.52 & 52.64 & 61.72 & 53.03 & 21,500 & 19,250 & Not reached & Not reached \\
\bottomrule
\end{tabular}
\end{table}

% Generated by scripts/build_artifacts.py.
\begin{table}[htbp]
\centering
\caption{Supplementary training-path ablations using the same seeds 42--44 in every row, LAST 40k. Entries are final-position accuracy (\%), mean $\pm$ sample SD. Input-SG stops gradients at all raw-update Q/K/V inputs of correction attention; the correction and its parameters remain active. Fixed uses scale 0.05. Upper decay reaches zero at 36k, so its ON and OFF endpoints coincide. Upper fixed/input-SG reuse the corresponding three-seed subset of Table~\ref{tab:upper-gradient}. Upper fixed is an instrumented repeat and lower fixed reuses completed runs; neither adds independent seeds.}
\label{tab:scaffold-primary}
\small
\setlength{\tabcolsep}{4pt}
\begin{tabular}{lrrr}
\toprule
Condition & ON: $L=40$ & ON: $L=80$ & OFF: $L=80$ \\
\midrule
Upper fixed & $99.35 \pm 0.50$ & $71.35 \pm 7.51$ & $71.09 \pm 7.77$ \\
Upper input-SG & $98.96 \pm 0.63$ & $68.68 \pm 12.26$ & $68.85 \pm 11.96$ \\
Upper decay & $99.41 \pm 0.49$ & $73.93 \pm 9.72$ & $73.93 \pm 9.72$ \\
Lower input-SG & $99.02 \pm 0.43$ & $66.28 \pm 14.05$ & $29.88 \pm 15.23$ \\
Lower fixed & $98.31 \pm 1.24$ & $62.60 \pm 12.92$ & $22.56 \pm 5.70$ \\
\bottomrule
\end{tabular}
\end{table}

% Generated by scripts/build_artifacts.py.
\begin{figure}[htbp]
\centering
\paperfigure{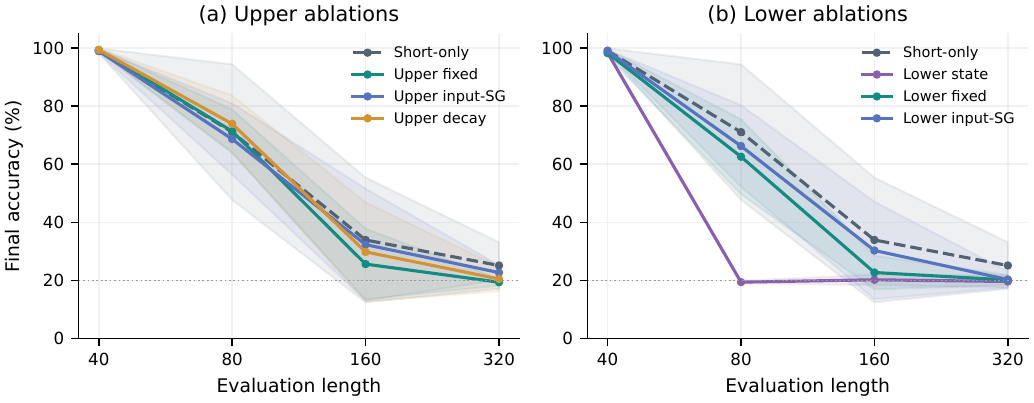}
\caption{Length sensitivity of the scaffold ablations at LAST (40k), using the same seeds 42--44 for every curve. Lines show means; bands show one sample SD clipped to [0,100]. The short-only, lower-state, and lower-fixed controls reuse completed runs. Upper fixed/input-SG use the corresponding three-seed subset of Table~\ref{tab:upper-gradient}. All settings degrade substantially beyond length 80. Upper decay has zero correction at evaluation; the other delta curves use their trained scale of 0.05. No configuration is selected using these extrapolation scores.}
\label{fig:scaffold-lengths}
\end{figure}

% Generated by scripts/build_artifacts.py.
\begin{table}[htbp]
\centering
\caption{Individual-seed scaffold results for the same seeds 42--44 in every condition. Accuracy (\%) uses the fixed LAST test set at $L=80$. The last column is the first ordinary $L=40$ validation final-position accuracy of at least 99\%, checked every 250 updates on 2,000 examples. It is a first crossing, not sustained attainment; ``Not reached'' refers to this validation threshold within 40k, not failure on every evaluation.}
\label{tab:scaffold-seeds}
\small
\setlength{\tabcolsep}{4pt}
\begin{tabular}{lrrrr}
\toprule
Condition & Seed & ON & OFF & First 99\% (step) \\
\midrule
Upper fixed & 42 & 79.88 & 79.88 & 29,500 \\
Upper fixed & 43 & 68.46 & 68.26 & 21,250 \\
Upper fixed & 44 & 65.72 & 65.14 & 28,500 \\
Upper input-SG & 42 & 82.03 & 81.84 & 33,000 \\
Upper input-SG & 43 & 66.11 & 66.41 & 32,750 \\
Upper input-SG & 44 & 57.91 & 58.30 & 33,250 \\
Upper decay & 42 & 82.52 & 82.52 & 30,750 \\
Upper decay & 43 & 75.88 & 75.88 & 21,250 \\
Upper decay & 44 & 63.38 & 63.38 & 31,500 \\
Lower input-SG & 42 & 79.69 & 47.46 & 32,500 \\
Lower input-SG & 43 & 67.48 & 21.58 & 31,250 \\
Lower input-SG & 44 & 51.66 & 20.61 & Not reached \\
\bottomrule
\end{tabular}
\end{table}

% Generated by scripts/build_artifacts.py.
\begin{table}[htbp]
\centering
\caption{Length and checkpoint sensitivity for the scaffold ablations, using the same seeds 42--44 in every row. Entries are final-position accuracy (\%), mean $\pm$ sample SD. BEST is selected solely by ordinary length-40 validation cross-entropy. Upper decay is already at zero scale at every selected BEST checkpoint.}
\label{tab:scaffold-sensitivity}
\footnotesize
\setlength{\tabcolsep}{4pt}
\begin{tabular}{lrrrr}
\toprule
Condition & LAST: $L=160$ & LAST: $L=320$ & BEST: $L=80$ & BEST: $L=160$ \\
\midrule
Upper fixed & $25.62 \pm 12.19$ & $19.37 \pm 1.17$ & $73.73 \pm 13.85$ & $30.47 \pm 17.64$ \\
Upper input-SG & $32.32 \pm 19.27$ & $22.69 \pm 2.52$ & $60.09 \pm 17.11$ & $33.43 \pm 20.58$ \\
Upper decay & $29.82 \pm 17.06$ & $20.51 \pm 4.32$ & $73.96 \pm 10.80$ & $30.76 \pm 18.09$ \\
Lower input-SG & $30.31 \pm 16.82$ & $20.31 \pm 2.76$ & $66.50 \pm 14.58$ & $28.78 \pm 14.96$ \\
\bottomrule
\end{tabular}
\end{table}

Table~\ref{tab:upper-gradient-seeds} gives all seven upper input-gradient
outcomes at both validation-accuracy thresholds. All seven seeds reach 95\%
in both conditions; the two seed-48 nonattainments at 99\% are reported
separately. The supplementary scaffold tables and length curves use
seeds 42--44 for every condition. Upper fixed/input-SG reuse this subset
of the seven-seed results to allow comparisons with upper decay and lower
input-SG on the same training seeds. The learning diagnostics also use
seeds 42--44.

\paragraph{Scheduled upper-only removal.}
Table~\ref{tab:scaffold-primary} includes upper decay, which finishes at
$73.93\pm9.72$\% at length 80 after 4,000 updates with zero correction.
This supports planned removal in the tested upper configuration.
The result differs from the all-layer decay mean
of 52.64\% but does not establish an optimal schedule. Lower input-SG also
depends on its correction within the training lengths: OFF reduces its
length-40 accuracy to $37.73\pm18.54$\%.

Every training run completes 40,000 updates. Instrumented fixed repeats
add diagnostic measurements without adding independent training seeds.
Frozen per-seed endpoint and diagnostic records
support all tables and curves. The source audit checks shared evaluation
fingerprints, correction bounds, completed budgets, and unchanged training
random-number streams. The upper-fixed and upper-decay model checkpoints and
data streams are identical at initialization and update 24,000.
The seven-seed upper input-SG audit verifies completed budgets,
matching evaluation fingerprints, and matching model/training settings.
Its raw metrics and source hashes are frozen alongside the other evidence.

For the supplementary scaffold experiments,
Table~\ref{tab:scaffold-seeds} makes the threshold definition and individual
outcomes explicit. The lower input-SG seed 44 does not cross 99\% on the
ordinary validation set during training; this does not imply that the other
metrics or independent final test set fail similarly.
Table~\ref{tab:scaffold-sensitivity} and Figure~\ref{fig:scaffold-lengths}
report length and checkpoint sensitivity for these supplementary experiments.
At length 80, the BEST means for upper decay and lower input-SG are
73.96\% and 66.50\%, respectively. All upper-decay BEST checkpoints
occur after the scale reaches zero.

As an additional carry-use check, suppressing recurrent phase inheritance
while retaining each run's scheduled correction gives length-80 means of
19.95\% for upper decay and 19.92\% for lower input-SG.
This evaluation-only intervention is distinct from disabling delta attention.
It supports reliance on recurrent carry at length 80 without identifying the
learned update algorithm. The data and artifact-generation checks are included
in the source bundle; no raw model checkpoints are required to rebuild figures.

\section{Correction magnitude and functional dependence during training}
\label{app:dynamics}
% Generated by scripts/build_artifacts.py.
\begin{figure}[tb]
\centering
\paperfigure{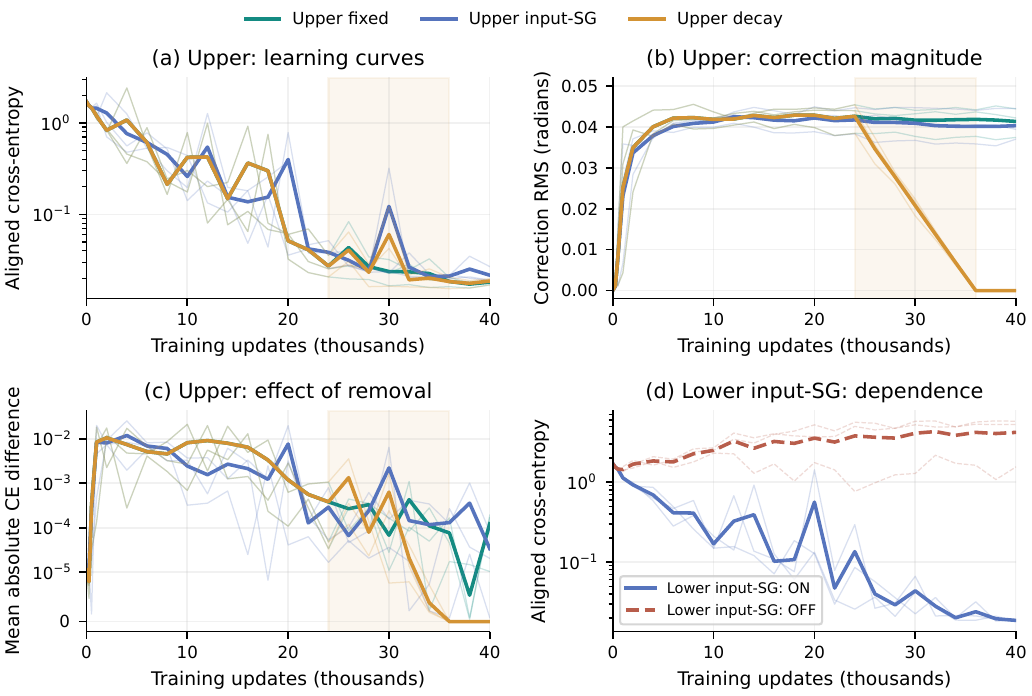}
\caption{Learning diagnostics on 512 fixed length-40 validation examples at 24 checkpoints. Bold lines average seeds 42--44; thin lines show individual seeds. (a) ON aligned cross-entropy. (b) Actual correction RMS, including the scheduled scale. (c) The per-seed absolute ON--OFF cross-entropy difference, averaged after taking the absolute value; the vertical scale is symmetric-log with a linear region below $10^{-5}$. (d) Lower input-SG remains strongly dependent on its forward correction. Shading marks the upper decay interval (24k--36k). Fixed-scale upper corrections remain nonzero while their effect on loss becomes small. Diagnostics do not select checkpoints or control training.}
\label{fig:scaffold-dynamics}
\end{figure}

Figure~\ref{fig:scaffold-dynamics} follows the correction during training.
On the fixed length-40 probes, upper-fixed correction RMS rises from 0.0251
radians at update 1,000 to 0.0423 at 8,000 and remains 0.0414 at 40,000.
It does not naturally decay to zero. Meanwhile, the mean absolute ON--OFF
aligned cross-entropy difference falls from 0.00840 at update 1,000 to
0.000123 at 40,000, with intermediate fluctuations. The upper correction can
remain numerically substantial relative to its 0.05-radian bound while having
little effect on the measured loss. A small absolute loss effect therefore
does not require the helper to shrink. This does not establish spontaneous
disappearance of the helper or that the uncorrected writes absorb its computation.

Lower input-SG provides a contrasting trajectory. At 40,000 its correction
RMS is only 0.0178 radians, smaller than the upper-fixed value, but removing
it raises the aligned probe cross-entropy from 0.0186 by 4.1795. Thus larger
removal damage does not imply a larger correction. The upper-decay correction
and ON--OFF gap become exactly zero at 36k by construction. These diagnostics
separate the measured amplitude, the effect of removal, and the imposed
schedule; none alone identifies the internal state semantics.

\end{document}